\documentclass[letterpaper, 10 pt, conference]{ieeeconf}

\IEEEoverridecommandlockouts

\usepackage{cite}
\usepackage{ulem}
\usepackage{soul}
\usepackage{acronym}
\usepackage{caption}
\usepackage{leftidx}
\usepackage{listings}
\usepackage{graphicx}
\usepackage{textcomp}
\usepackage{multirow}
\usepackage{booktabs}
\usepackage{colortbl}
\usepackage{subcaption}
\usepackage{algorithm}
\usepackage[noend]{algpseudocode}
\usepackage{amsmath,amssymb,amsfonts}

\usepackage{booktabs}
\usepackage{array}
\usepackage{savesym}
\savesymbol{checkmark}
\usepackage{dingbat}
\usepackage{dingbat}
\usepackage{bbding}
\usepackage{colortbl}
\usepackage[table]{xcolor}
\usepackage{lipsum}
\usepackage{censor}
\usepackage{comment}

\usepackage[utf8]{inputenc}
\usepackage{pgfplots}
\DeclareUnicodeCharacter{2212}{−}
\usepgfplotslibrary{groupplots,dateplot}
\usetikzlibrary{patterns,shapes.arrows}
\pgfplotsset{compat=newest}

\usepackage{hyperref}
\usepackage{cleveref}

\acrodef{MR}{Mixed Reality}
\acrodef{AR}{Augmented Reality}
\acrodef{HMD}{Head-Mounted Display}
\acrodef{HRI}{Human-Robot Interaction}
\acrodef{SLAM}{Simultaneous Localization and Mapping}

\definecolor{red}{HTML}{fd8f8f}
\definecolor{greend}{HTML}{57e377}
\definecolor{greenl}{HTML}{b8fb8a}

\colorlet{red}{red!50}
\colorlet{greenl}{greenl!50}
\colorlet{greend}{greend!50}

\title{\LARGE \bf Human Interaction for Collaborative Semantic SLAM using Extended Reality}

\author{
     Laura Ribeiro$^{1}$, Muhammad Shaheer$^{1}$, Miguel Fernandez-Cortizas$^{1}$, Ali Tourani$^{1}$ \\ Holger Voos$^{1,2}$, and Jose Luis Sanchez-Lopez$^{1}$
    \thanks{$^{1}$Automation and Robotics Research Group, Interdisciplinary Centre for Security, Reliability, and Trust (SnT) and $^{2}$Faculty of Science, Technology, and Medicine. Both from the University of Luxembourg, Luxembourg.
    \tt{\small{\{laura.ribeiro, muhammad.shaheer, miguel.fernandez, ali.tourani, holger.voos, joseluis.sanchezlopez\}}@uni.lu}}
    \thanks{This research was funded, in whole or in part, by the Luxembourg National Research Fund (FNR) under the MR-Cobot Project (Ref. 18883697/MR-Cobot) and the DEUS Project (Ref. C22/IS/17387634/DEUS). Also by the Institute of Advanced Studies (IAS)/ “Audacity” grant (project TRANSCEND - 2021).} 
    \thanks{The authors gratefully acknowledge the students Gregorio Orlando and Maxime Capdouze for their valuable contributions to the technical aspects of the project setup.}
    \thanks{*For the purpose of open access, and in fulfillment of the obligations arising from the grant agreement, the author has applied a Creative Commons Attribution 4.0 International (CC BY 4.0) license to any  Author Accepted Manuscript version arising from this submission.}}

\begin{document}

\maketitle
\thispagestyle{empty}
\pagestyle{empty}

\begin{abstract}
Semantic SLAM (Simultaneous Localization and Mapping) systems enrich robot maps with structural and semantic information, enabling robots to operate more effectively in complex environments. However, these systems struggle in real-world scenarios with occlusions, incomplete data, or ambiguous geometries, as they cannot fully leverage the higher-level spatial and semantic knowledge humans naturally apply. We introduce HICS-SLAM, a Human-in-the-Loop semantic SLAM framework that uses a shared extended reality environment for real-time collaboration. The system allows human operators to directly interact with and visualize the robot's 3D scene graph, and add high-level semantic concepts (e.g., rooms or structural entities) into the mapping process. We propose a graph-based semantic fusion methodology that integrates these human interventions with robot perception, enabling scalable collaboration for enhanced situational awareness. Experimental evaluations on real-world construction site datasets demonstrate improvements in room detection accuracy, map precision, and semantic completeness compared to automated baselines, demonstrating both the effectiveness of the approach and its potential for future extensions.
\end{abstract}
\section{Introduction}
\label{sec_intro}

Simultaneous Localization and Mapping (SLAM) constructs environmental maps while estimating robot position \cite{nashed_human---loop_2018}. Traditional approaches focus on geometric representations, while Semantic SLAM enriches maps with structural and semantic information, enabling systems to identify objects, entities, and their associations beyond geometry \cite{hughes_hydra_2022, bavle_situational_2022}. However, semantic mapping remains sensitive to noise and misclassified elements, which propagate errors and reduce map robustness \cite{muravyev_evaluation_2023}. These limitations affect applications requiring accurate spatial understanding, including search-and-rescue operations, industrial assembly \cite{chang_survey_2024, guerrero_augmented_2022}, and construction sites. To address these challenges, Human-in-the-Loop (HitL) strategies incorporate human knowledge into the mapping process.

\begin{figure}[t]
    \centering
    \includegraphics[width=1\linewidth]{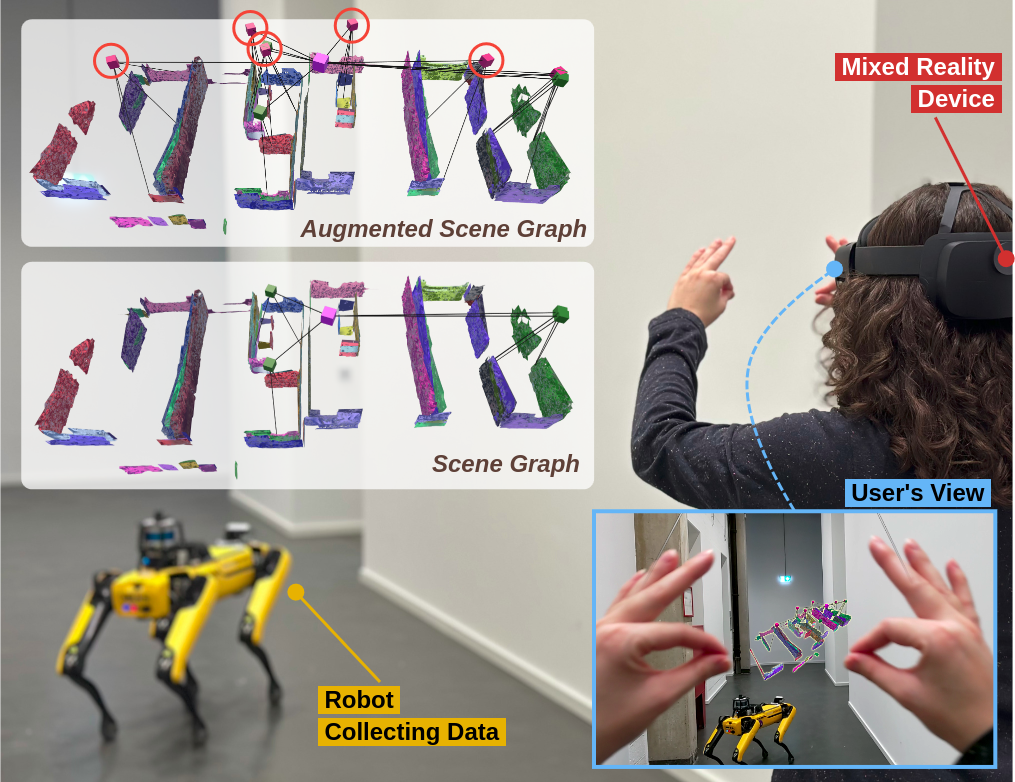}
    \caption{The overall idea behind HICS-SLAM: a human interacts with the map through the proposed extended reality application in a virtual world, adding new rooms (characterized by red circles) to the generated 3D scene graph and updating them in real time. These modifications propagate through the SLAM algorithm, enhancing mapping accuracy and localization performance simultaneously.}
    \label{fig:overall}
\end{figure}

HitL methodologies enable interactive correction of trajectory estimation and semantic labeling during SLAM operation \cite{sidaoui_-slam_2019}. Recent works, like HAC-SLAM \cite{sayour_hac-slam_2024}, introduced trajectory corrections in the SLAM, while HSS-SLAM \cite{li_hss-slam_2024} used superquadric representations for map refinement. However, the methods could be improved using high-level semantic editing, since HAC-SLAM lacks semantic integration into scene graphs, and HSS-SLAM requires manual parameter adjustments that are not intuitive for non-experts and cannot support room-scale semantic editing.

We introduce HICS-SLAM, a new approach that combines HitL features with Semantic SLAM through a virtual environment. HICS-SLAM enables high-level semantic user operations, such as creating new rooms in the scene graph, through an Extended Reality (XR) shared environment. 

XR interfaces provide intuitive means to inspect reconstructed environments and perform corrections through natural hand gestures \cite{m2023enhancing, suzuki_augmented_2022}, offering advantages over 2D interfaces through spatially grounded interactions that leverage human spatial cognition \cite{tang_real-time_2023}. HICS-SLAM uses scene graphs, which represent environments hierarchically as nodes and edges, allowing SLAM systems to optimize objects and rooms \cite{hughes_hydra_2022}. This hierarchical representation enables interventions at the graph level, allowing corrections to propagate across related elements while maintaining global semantic and geometric consistency \cite{bavle_s-graphs_2025}.

Combining XR with a 3D scene graph, HICS-SLAM allows users to inspect reconstructed environments and define semantic structures through hand gestures, which are integrated into the S-Graphs 2.0 \cite{bavle_s-graphs_2025} backend and propagate through the SLAM system. \Cref{fig:overall} shows a human operator utilizing the HICS-SLAM system. This workflow supports real-time data collection, SLAM processing, and interactive human refinement, shifting manual corrections from post-processing to real-time while improving map quality and localization consistency.

The main contributions of this work are:
\begin{itemize}
    \item A collaborative SLAM framework that enables intuitive human intervention through XR interfaces, allowing direct manipulation of 3D scene graphs for spatial modeling.
    \item An online integration method that incorporates human semantic knowledge into the SLAM back-end optimization, improving mapping quality through high-level corrections.
    \item Real-world experimental validation demonstrating the effectiveness of human-guided semantic corrections for improving SLAM applications.
\end{itemize}
\section{Related Works}
\label{sec_related}

HitL strategies have become powerful approaches for combining human interaction with automated mapping and navigation systems \cite{nashed_human---loop_2018}. Early methods employed factor graph extensions to allow joint optimization of trajectories and maps \cite{nashed_human---loop_2018}, while subsequent systems provided mechanisms for interactive trajectory correction, such as HAC-SLAM \cite{sayour_hac-slam_2024}, and semantic-assisted mapping, as in A-SLAM \cite{sidaoui_-slam_2019}. 

More recent frameworks incorporate semantic representations to facilitate higher-level interventions, including HSS-SLAM, which models environments with superquadrics \cite{li_hss-slam_2024}, Collaborative Human Augmented SLAM, enabling multi-user semantic collaboration \cite{sidaoui_collaborative_2019}, and approaches that automatically infer semantic-relational structures from factor graphs \cite{learning_jaslam}. Evaluations of topological mapping methods in indoor environments \cite{muravyev_evaluation_2023} highlight challenges that remain for automated systems, motivating human intervention approaches.

Extended Reality methodologies have enhanced human-robot collaboration by providing intuitive interfaces for map inspection and manipulation. Direct SLAM applications include Interactive 3D Graph SLAM \cite{koide_interactive_2021}, Mixed Reality (MR) Robot Control \cite{ostanin_interactive_2019}, and 3D MR interfaces for human-robot teaming \cite{chen_3d_2024}, which enable real-time 3D visualization and map corrections. Related collaborative navigation approaches include intuitive human-drone navigation in unknown environments through mixed reality \cite{drone_mixed_slam} and spatial-assisted human-drone collaborative frameworks \cite{drone_mixedreality}. Broader applications demonstrate XR potential in robotics: HoloDesigner \cite{dan_holodesigner_2021} for on-site design, mixed-reality teleoperation for high-level control of legged-manipulator robots \cite{ulloa_mixed-reality_2022}, Boston Dynamics Spot integration with Microsoft HoloLens 2 (MR Device) \cite{hofe_demo_2023} for robot teleoperation, MR integration with Building Information Modeling (BIM) \cite{bahri_efficient_2019}, digital twin frameworks for interactive building operations \cite{ma_digital_2025, tang_real-time_2023}, and comprehensive XR approaches for enhanced human-robot collaboration \cite{karpichev_extended_2024}.

Despite these advances, most existing HitL approaches either operate at low-level geometric representations or provide high-level interaction without incorporating semantic information, making manual intervention time-consuming, prone to inaccuracies, and less effective in complex or large-scale environments. Our approach bridges this gap by enabling high-level semantic map interaction in XR, combining intuitive scene manipulation with semantic awareness to support scalable human-robot collaboration.
\section{System Overview}
\label{sec_proposed}

\begin{figure*}[t]
\vspace{0.8em}
    \centering
    \includegraphics[width=\linewidth]{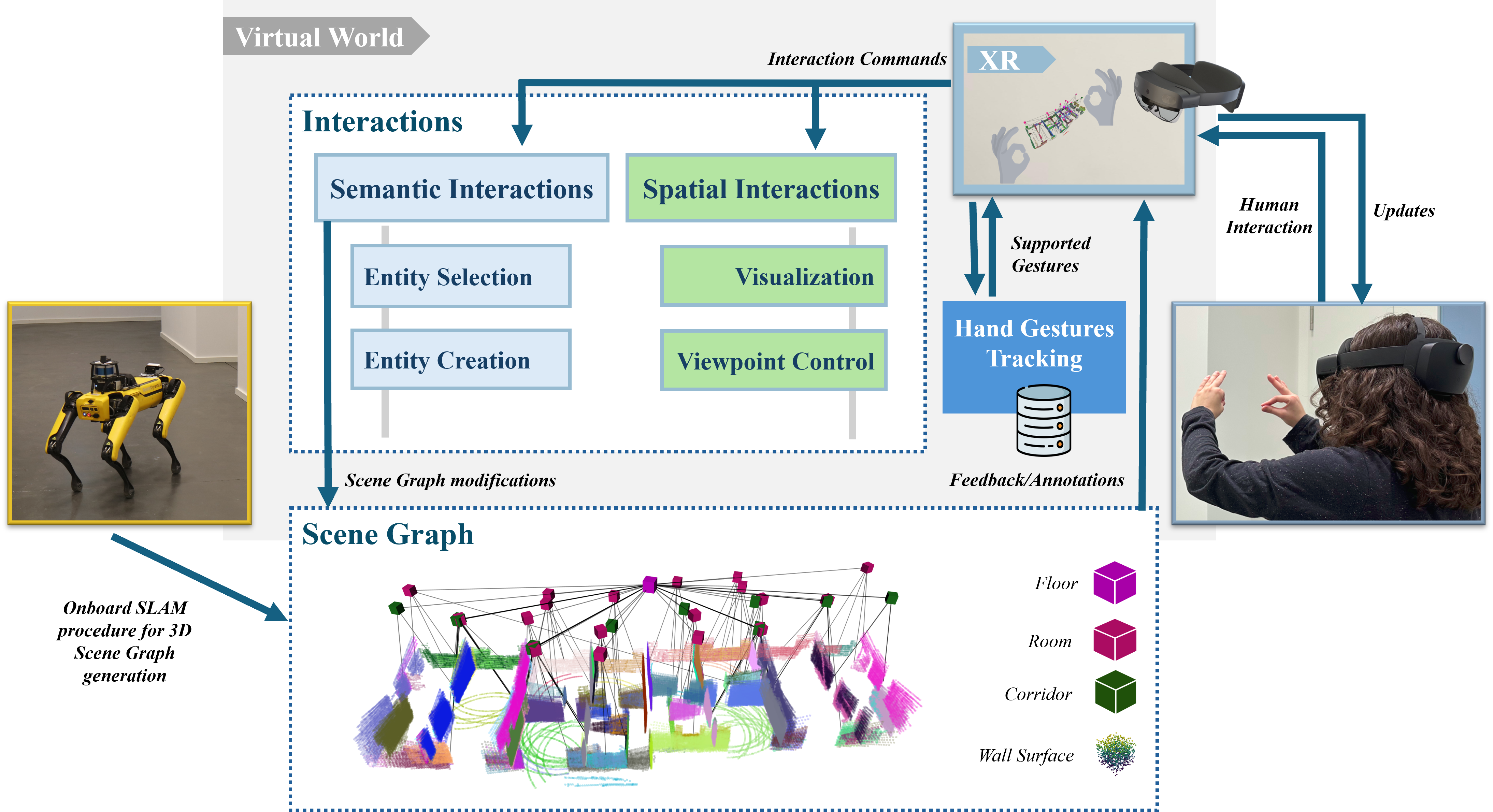}
    \caption{System architecture overview. The proposed method enables real-time collaboration between the robot and human through a shared virtual environment. The robot operates in the real world while generating a hierarchical scene graph representation, and humans interact with this virtual representation using the HoloLens 2 device and hand gestures to enhance semantic understanding, creating a bidirectional knowledge exchange between real and virtual worlds.}
    \label{fig:architecture}
\end{figure*}

\subsection{Virtual World, Real World: Collaborative Mapping}

Our system enables collaborative mapping where a robot operates in the physical environment and collaborates with a human operator through a shared virtual representation of the robot's situational awareness.
The robot collects sensor data from its environment, constructing a hierarchical world representation that encompasses geometric primitives (planes, keyframes) and semantic concepts (rooms, walls, objects).

This knowledge is accessible to the human operator through an MR device that renders an interactive virtual environment.
The human operator augments the robot's semantic understanding by defining spatial relationships using natural hand gestures.
Hence, this bidirectional knowledge exchange leverages the complementary capabilities of robotic sensing and human spatial cognition to enhance Semantic SLAM in real-time, as shown in \Cref{fig:architecture}.

In the \textit{Real World}, sensor data are processed through the SLAM algorithm to generate the robot's geometric and semantic scene understanding. On the other hand, in the \textit{Virtual World}, the human visualizes this understanding as an interactive view of the shared model, where the human operator provides semantic augmentations.

\subsection{Real World}

Sensory data collected by the robot is processed through semantic SLAM algorithms to construct environmental knowledge. This processing generates both geometric and semantic representations, organizing them in a hierarchical structure where high-level semantic elements (rooms) connect to low-level geometric entities (planes).

This hierarchical organization enables geometric primitives, such as planes, to be aggregated into higher-level semantic entities, such as rooms, through graph-based connections during optimization. The system maintains consistent spatial relationships while incorporating meaningful environmental structures that facilitate human-robot mutual understanding. The graph-based representation ensures efficient updates and scalable performance as environmental complexity increases.

For our implementation, the project uses S-Graphs 2.0 \cite{bavle_s-graphs_2025} as the underlying SLAM algorithm. S-Graphs is a metric-relational semantic SLAM that organizes the environment hierarchically and semantically in a graph. Its key elements include:
\begin{itemize}
    \item \textbf{Keyframes:} Robot sensor poses capturing geometry and observations within each room.
    \item \textbf{Walls and Planes:} Structural components are linked to their respective rooms.
    \item \textbf{Rooms:} Child nodes under floors, containing spatial boundaries and semantic labels.
    \item \textbf{Floors:} Top-level nodes representing the levels, where the algorithm handles multifloors. 
\end{itemize}
Through the framework, humans provide semantic information to S-Graphs, with \textit{rooms} serving as a key example. In this context, a room represents a higher-level semantic concept that groups multiple \textit{planes}.

\subsection{Virtual World}

The Virtual World, or virtual environment, serves as a bridge between human operators and the robot's knowledge representation, rendering the hierarchical scene graph in an interactive 3D map. It receives real-time sensor data updates and converts raw point clouds into plane structures, which are then represented interactively for the user through MR devices. 

Human operators use XR visualization and hand gestures to contribute information and correct SLAM errors within this virtual environment. The system supports two interaction modalities: (i) semantic interactions and (ii) spatial interactions.
Spatial interactions allow operators to navigate the virtual environment through zoom, rotation, and manipulation gestures, facilitating detailed inspection of the robot's mapping state. Semantic interactions enable operators to select geometric primitives and construct higher-level semantic concepts that the SLAM algorithm missed or misinterpreted. For instance, when the algorithm fails to recognize room boundaries, operators can manually combine four wall surfaces, here as four planes, to define the complete room structure.

Bidirectional communication with the Real World integrates human semantic contributions into the robot's knowledge base while propagating robot discoveries to update the virtual representation, creating a collaborative SLAM framework that augments autonomous mapping with human semantic reasoning.

\subsection{Architecture Implementation}
The HICS-SLAM implementation builds a virtual world that reflects the robot’s situational awareness through S-Graphs operating in the real world. We describe how this semantic representation affects and extends the underlying SLAM system.

\subsubsection{\textbf{3D Scene Graphs}}
The application reconstructs the S-Graphs hierarchy, including floors, rooms, walls, and keyframes, reflecting the robot’s internal understanding of the environment.
The virtual world enables intuitive user interaction with natural hand gestures for navigating and interacting with entities from the hierarchical graph.
The reconstruction procedure ensures the virtual world faithfully reflects the robot's understanding and perception of the real world.

In the HICS-SLAM, creating a room from four planes introduces an edge in the hierarchical factor graph. This edge is assigned higher precision weights in the information matrix to encode confidence in the human operator’s inputs. Each room enforces consistency between the room vertex and its four surrounding planar boundaries, constraining the room's center ($p_{room}$) to coincide with the geometric center of the enclosed space by the planes.
The optimization phase in the factor graph aims to reduce the distance between these two measurements.
The residual error ($\mathbf{e}$) is calculated as:
$$\mathbf{e} = \mathbf{p}_{room} - (\mathbf{v}_x + \mathbf{v}_y)$$
where the displacement vectors ($\mathbf{v}_x$ and $\mathbf{v}_y$) represent the estimated geometric center of the room.
These vectors are calculated from the distances $d_x$$,_y$ and normals $\mathbf{n}_i$ of the surrounding planes:
\begin{equation*}
    \begin{cases}
    \mathbf{v}_x = d_x\mathbf{n}_x + \frac{1}{2}(d_2 \mathbf{n}_2 + d_1 \mathbf{n}_1), & \text{(x-axis)} \\
    \mathbf{v}_y = d_y\mathbf{n}_y + \frac{1}{2}(d_4 \mathbf{n}_4 + d_3 \mathbf{n}_3), & \text{(y-axis)}
    \end{cases}
\end{equation*}
Here, $d_i$ is the distance component from the plane equation of each plane constraining the room's center, $\mathbf{n}_i$ is the unit normal vector pointing inward to the room, and $d_i\mathbf{n}_i$ is a reference plane with the smallest $d_i$ component in its axis direction.
The pairs $(d_1, d_2)$ and $(d_3, d_4)$ correspond to opposing planes that define the room's rectangular bounds in the horizontal plane. 

\subsubsection{\textbf{Interactions}}
\begin{figure}[t]
\vspace{0.8em}
    \centering
    \subfloat[\centering Zoom IN or OUT]{\includegraphics[width=4cm]{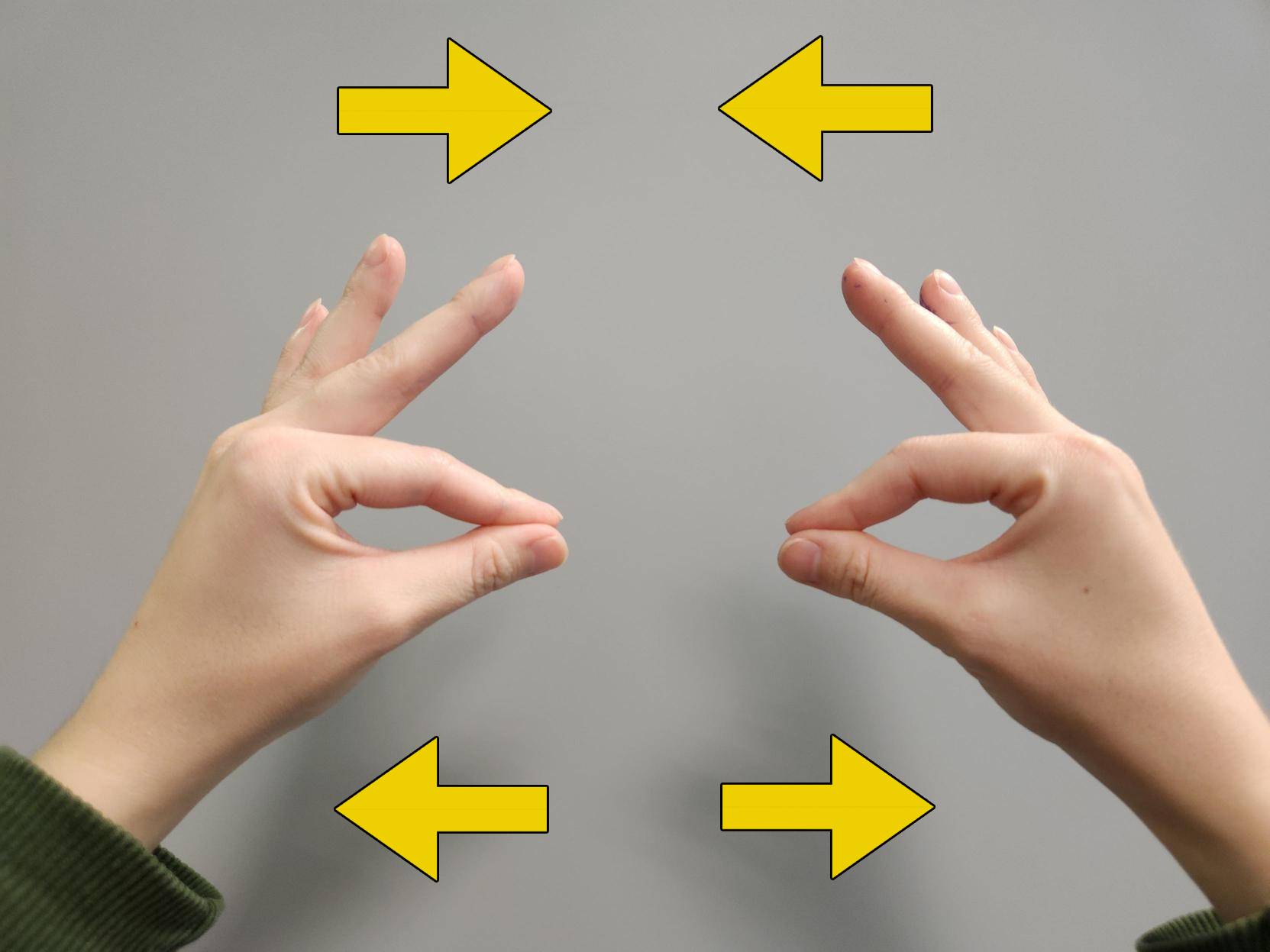}}
    \hspace{0.8em}
    \subfloat[\centering Move]{\includegraphics[width=4cm]{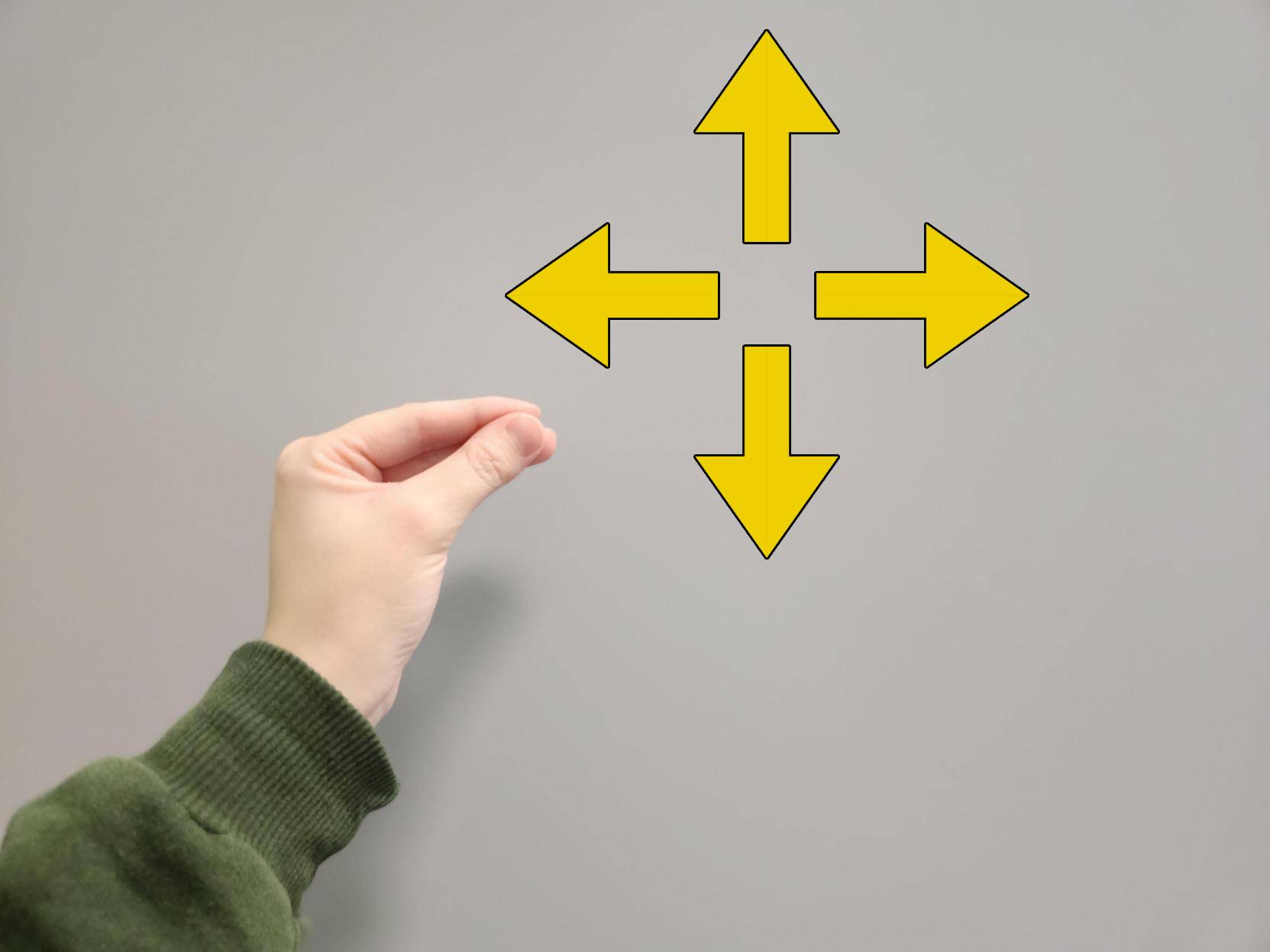}}
    \vspace{0.8em}
    \subfloat[\centering Rotate]{\includegraphics[width=4cm]{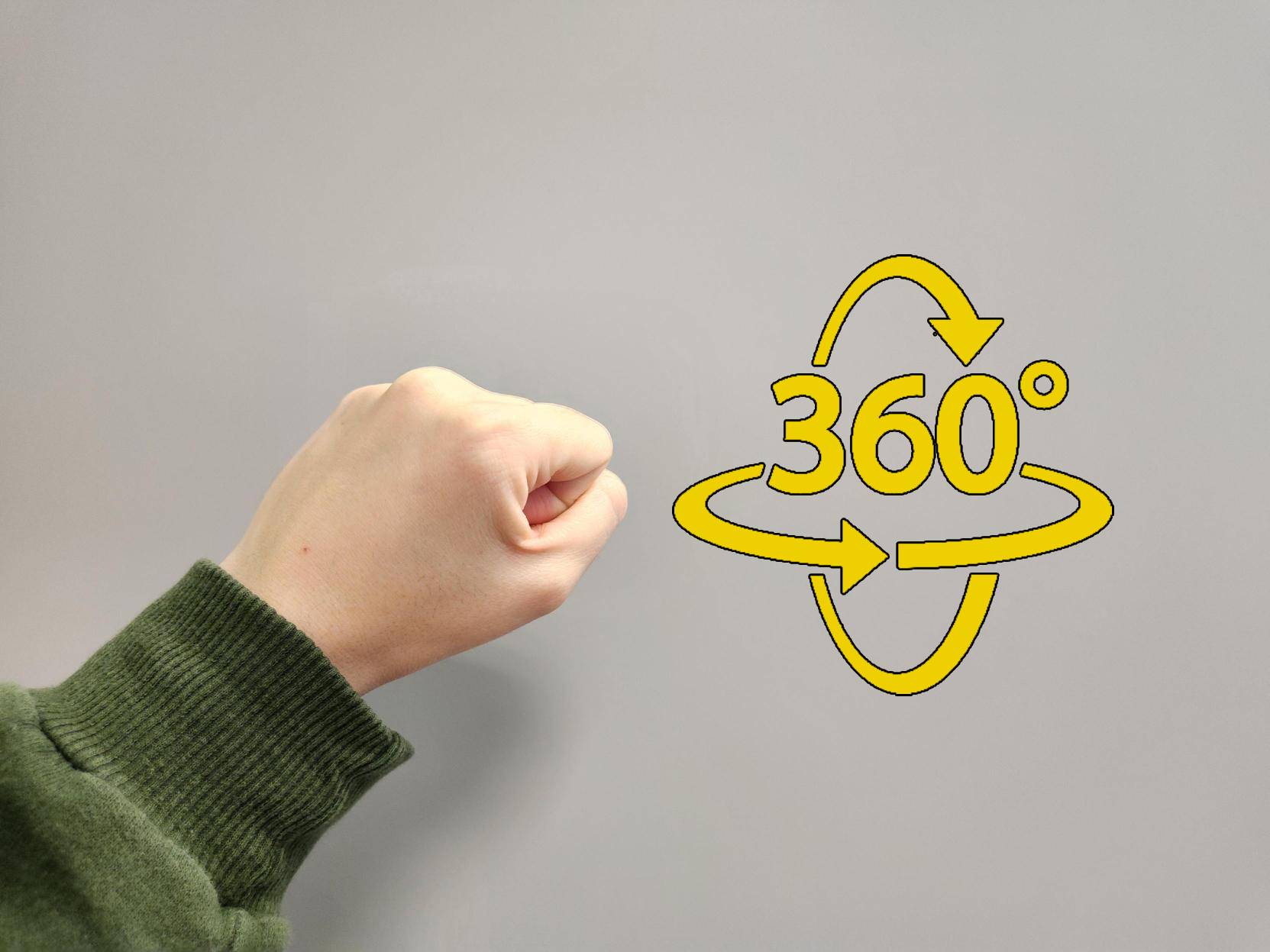}}
    \hspace{0.8em}
    \subfloat[\centering Pinch]{\includegraphics[width=4cm]{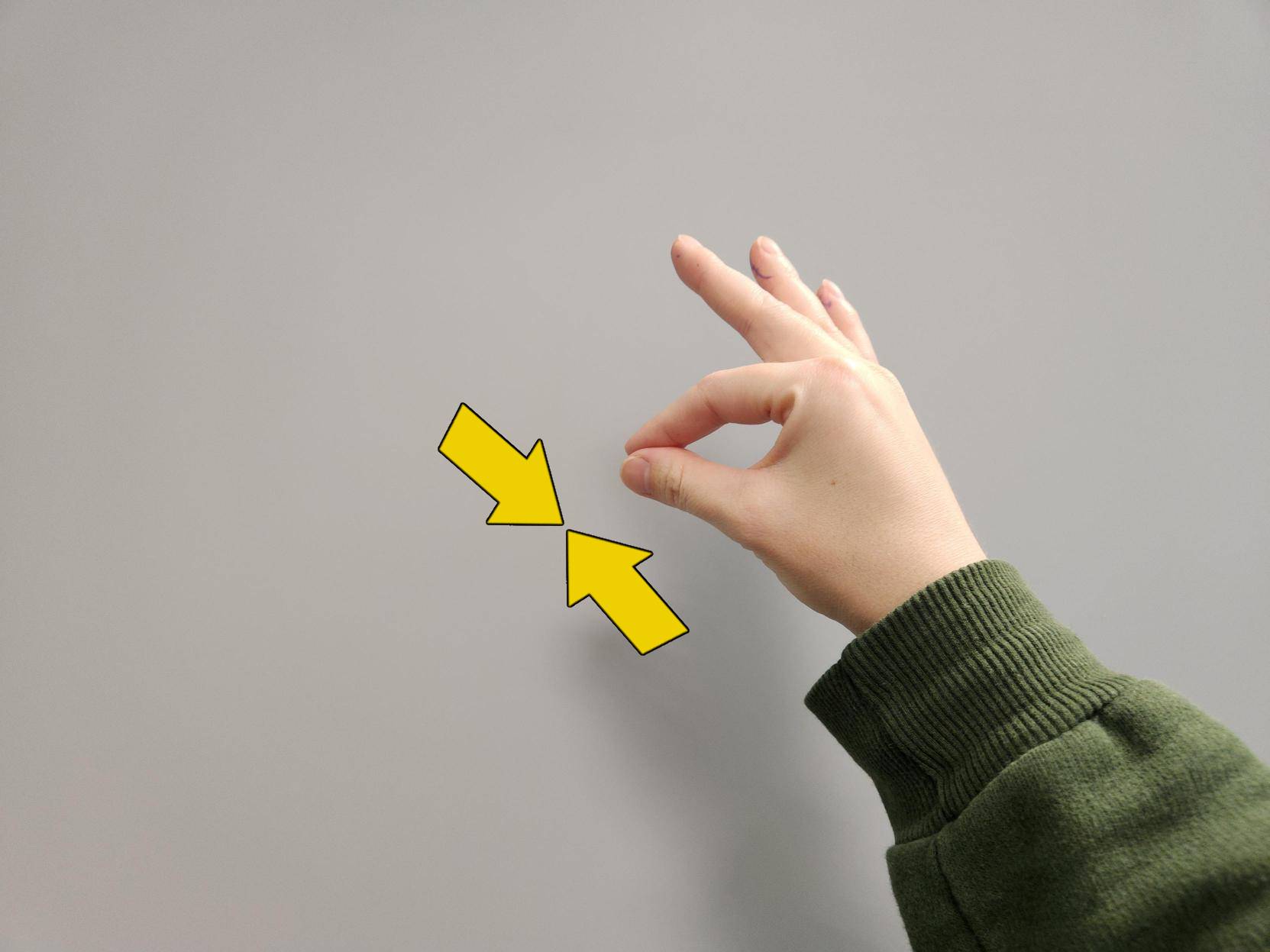}}
    \caption{Supported hand gestures for semantic and spatial interactions in the virtual world.
    (a) The two-handed pinch gesture zooms in and out of specific regions in the virtual world.
    (b) The left-hand pinch gesture moves the virtual representation around the x-axis and y-axis.
    (c) The grab gesture rotates the virtual representation around the x-axis, y-axis, and z-axis with the right or left hand.
    (d) The right-hand pinch gesture to select the entities in the virtual world.}
    \label{fig:human_interaction}
\end{figure}

In HICS-SLAM, we implemented natural hand gestures, as shown in \Cref{fig:human_interaction}, allowing human operators to navigate the hierarchical map and move freely within the virtual world. The map representation can be repositioned at any location the operator prefers to inspect the robot's situational awareness.


\section{Experimental Results}
\label{sec_evaluation}

\subsection{Evaluation Setup}

The XR interface was developed in Unity and tested using Microsoft HoloLens 2 (MR device), while remaining compatible with other devices, such as Meta Quest. The system communicates with the SLAM back-end via ROS 2.

We validate our approach in diverse real indoor environments datasets, ranging from university buildings to active construction sites. As in \cite{bavle_s-graphs_2025}, the datasets were recorded using a legged robot equipped with either a Velodyne VLP-16 or an Ouster OS-1 64 3D LiDAR. 

For the evaluation, we considered datasets collected from both real and simulated environments. \textit{C1F1} and \textit{C1F2} were obtained on two distinct floors of a compact construction site, while \textit{C2F0}, \textit{C2F1}, and \textit{C2F2} correspond to three floors of a larger ongoing project that integrates four individual houses. In turn, \textit{C3F1} and \textit{C3F2} represent two floors of a structure formed by the combination of two houses. Complementing these datasets, \textit{SC1F1} and \textit{SC1F2} were generated from 3D meshes of architectural plans, capturing two floors of an actual building. Finally, \textit{SE1}, \textit{SE2}, and \textit{SE3} are synthetic datasets specifically designed to simulate typical indoor environments.

\begin{figure}[t]
\vspace{0.8em}
    \centering
    \includegraphics[width=0.95\linewidth]{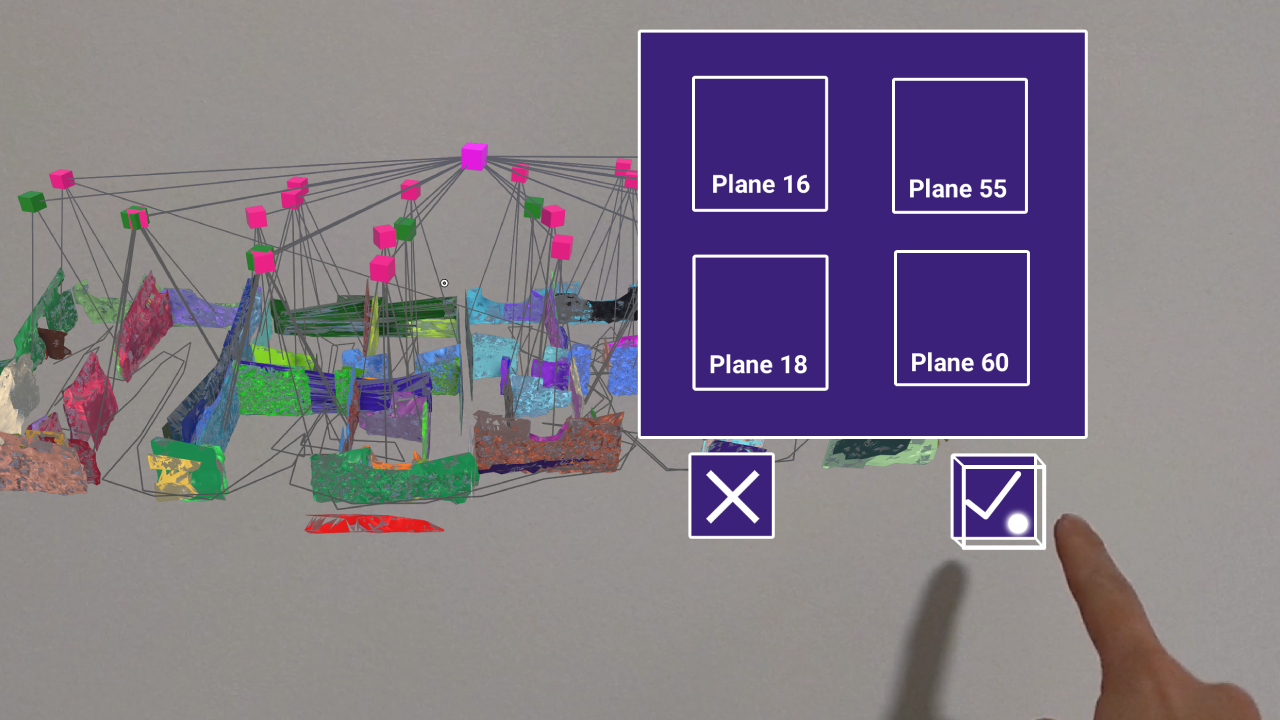}
    \caption{Demonstration of creating a room via human interaction using the proposed approach. The user confirms to send the four selected planes to the scene graph.}
    \label{fig:selectionplanes}
\end{figure}

The functionality to integrate the semantic information of \textit{rooms} into the system was implemented. \Cref{fig:selectionplanes} illustrates the human operator confirming the integration of the new room in S-Graphs.

\subsection{Evaluation Methodology} 
We evaluate our approach along two complementary aspects: the geometric accuracy of the reconstructed 3D maps and the structural expressiveness of the generated scene graphs in terms of room detection. 

These evaluations are conducted across both simulated environments and real-world datasets. Human operators interact with the reconstructed map via HoloLens 2 through a custom Unity application. Upon creation, new rooms are transmitted to S-Graphs via ROS topic, without altering the original wall geometry.

Our contributions aim to improve both map accuracy and trajectory estimation. To evaluate mapping and localization performance, we utilize the \textbf{Root Mean Square Error (RMSE)} and \textbf{ Absolute Trajectory Error (ATE)}. 

Room detection performance is quantified using standard classification metrics: \textbf{Precision}, \textbf{Recall}, and \textbf{F1-Score}, by comparing detected rooms against ground truth annotations established through manual inspection of each test environment. These metrics provide complementary insights: precision measures the accuracy of room detections (minimizing false positives), recall assesses the completeness of room identification (minimizing false negatives), and F1-score provides a balanced evaluation of overall detection performance. Additionally, the effectiveness of this interactive room definition process can also be evaluated through qualitative analysis, as detailed in Qualitative Results \ref{subsec:qualitative_results}.


\subsection{Results}

\begin{table}[t]
\vspace{0.8em}
\centering
\renewcommand{\arraystretch}{1.1} 
\setlength{\tabcolsep}{4pt}       
\scriptsize
\caption{Point cloud Root Mean Square Error (RMSE) values (in $\mathrm{cm}$) for the ``real" scenarios. The best results are \textbf{boldfaced}.}
\begin{tabular}{c|ccccccc|c}
\toprule
\multirow{2}{*}{\textbf{Method}} & \multicolumn{7}{c|}{\textbf{Point Cloud RMSE [$\mathrm{cm}$] $\downarrow$}} & \multirow{2}{*}{\textbf{Avg}} \\
\cmidrule(lr){2-8}
 & \textit{C1F1} & \textit{C1F2} & \textit{C2F0} & \textit{C2F1} & \textit{C2F2} & \textit{C3F1} & \textit{C3F2} \\
\midrule
S-Graphs\cite{bavle_s-graphs_2025}     & 32.87 & 19.22 & \textbf{13.17} & 21.48 & 17.96 & 25.14 & 18.93 & 19.22 \\
\midrule
Ours & \textbf{32.47} & \textbf{19.15} & 13.39 & \textbf{21.42} & \textbf{17.73} & \textbf{25.10} & \textbf{18.87} & \textbf{19.15} \\
\bottomrule
\end{tabular}
\label{tab:rmse_real_data}
\end{table}


\begin{table}[t]
\vspace{0.8em}
\centering
\renewcommand{\arraystretch}{1.1} 
\setlength{\tabcolsep}{4pt}       
\scriptsize
\caption{Absolute Trajectory Error (ATE) measurements for the ``simulated" scenarios. The best results are \textbf{boldfaced}.}
\begin{tabular}{c|ccccc|c}
\toprule
\multirow{2}{*}{\textbf{Method}} & \multicolumn{5}{c|}{\textbf{ATE [m \ensuremath{\times} 10$^{-2}$]$\downarrow$}} & \multirow{2}{*}{\textbf{Avg}} \\
\cmidrule(lr){2-6}
 & \textit{SE1} & \textit{SE2} & \textit{SE3} & \textit{SC1F1} & \textit{SC1F2}\\
\midrule
S-Graphs\cite{bavle_s-graphs_2025} & 5.07 & \textbf{1.35} & 2.42 & 2.98 & \textbf{8.29} & 2.99 \\
\midrule
Ours & \textbf{1.15} & 1.67 & \textbf{2.30} & \textbf{2.48} & \textbf{8.29} & \textbf{2.31}\\
\bottomrule
\end{tabular}
\label{tab:ate_simu_data}
\end{table}


\begin{table*}[htbp]
\vspace{0.8em}
\centering
\caption{Comparison of the performance metrics (Precision, Recall, and F1-Score) for room detection between our method and \textit{S-Graphs} 2.0~\cite{bavle_s-graphs_2025}, across different real-world datasets.}
\label{tab:room_detection_comparison}
\begin{tabular}{l|l|ccccccc|c}
\toprule
\multirow{2}{*}{\textbf{Method}} & \multirow{2}{*}{\textbf{Metrics}} & \multicolumn{7}{c|}{\textbf{Dataset}} & \multirow{2}{*}{\textbf{Avg}} \\
\cmidrule(lr){3-9}
 & & \textit{C1F1} & \textit{C1F2} & \textit{C2F0} & \textit{C2F1} & \textit{C2F2} & \textit{C3F1} & \textit{C3F2} & \\
\midrule
\multirow{3}{*}{S-Graphs\cite{bavle_s-graphs_2025}} & Precision $\uparrow$ & 1.00 & 1.00 & 1.00 & 0.00 & 1.00 & 0.00 & 1.00 & 0.71 $\pm$ 0.45 \\
 & Recall $\uparrow$ & 0.33 & 0.63 & 0.67 & 0.00 & 0.58 & 0.00 & 0.43 & 0.38 $\pm$ 0.27 \\
 & F1-Score $\uparrow$  & 0.50 & 0.77 & 0.80 & 0.00 & 0.74 & 0.00 & 0.60 & 0.49 $\pm$ 0.32 \\
\midrule
\multirow{3}{*}{\textit{Ours}} & Precision $\uparrow$  & 1.00 & 1.00 & 1.00 & 1.00 & 1.00 & 1.00 & 1.00 & 1.00 $\pm$ 0.00 \\
 & Recall $\uparrow$ & 1.00 & 1.00 & 0.83 & 1.00 & 1.00 & 0.91 & 0.93 & 0.95 $\pm$ 0.06 \\
 & F1-Score $\uparrow$ & 1.00 & 1.00 & 0.91 & 1.00 & 1.00 & 0.95 & 0.96 & 0.97 $\pm$ 0.03 \\
\bottomrule
\end{tabular}
\end{table*}


\textbf{Root Mean Square Error:} Table~\ref{tab:rmse_real_data} presents the RMSE evaluation of point cloud reconstructions across seven real-world datasets. HICS-SLAM demonstrates reliable geometric accuracy while incorporating human semantic knowledge, achieving small but consistent improvements in 6 out of 7 test scenarios, with enhancements of up to 1.23\% in scenario \textit{C1F1} and 1.21\% in \textit{C2F2}.

\textbf{Absolute Trajectory Error:} Table~\ref{tab:ate_simu_data} summarizes robot localization accuracy on simulated datasets. Our method demonstrates superior trajectory estimation performance, with trajectory errors reduced by up to 77.3\% in SE1, 17.1\% in SC1F1, and 5.0\% in SE3 compared to S-Graphs, achieving an overall improvement of 22.7\% in average ATE. 

\textbf{Performance Metrics for Room Detection:} Table~\ref{tab:room_detection_comparison} summarizes the results of room detection across the datasets. The baseline S-Graphs achieve consistently high precision but suffer from limited recall, often missing rooms that are not fully captured in the robot trajectory. 
In scenarios such as \textit{C2F0}, \textit{C3F1}, and \textit{C3F2}, some rooms cannot be fully created because the available point clouds do not capture the final planes. While this limitation affects both methods, our collaborative semantic SLAM approach mitigates it by integrating human knowledge and partial cues, allowing the reconstruction of rooms that would otherwise remain undetected. As a result, recall increases from an average of 38\% with S-Graphs to 95\% with our method, and F1-Score rises from 49\% to 97\%, leading to a more complete semantic representation of the environment even for partially observed or occluded rooms.

\subsection{Qualitative Results}
\label{subsec:qualitative_results}

\begin{figure*}[!t]
  \centering
  \begin{subfigure}{0.3\textwidth}
    \centering
    \includegraphics[width=\linewidth]{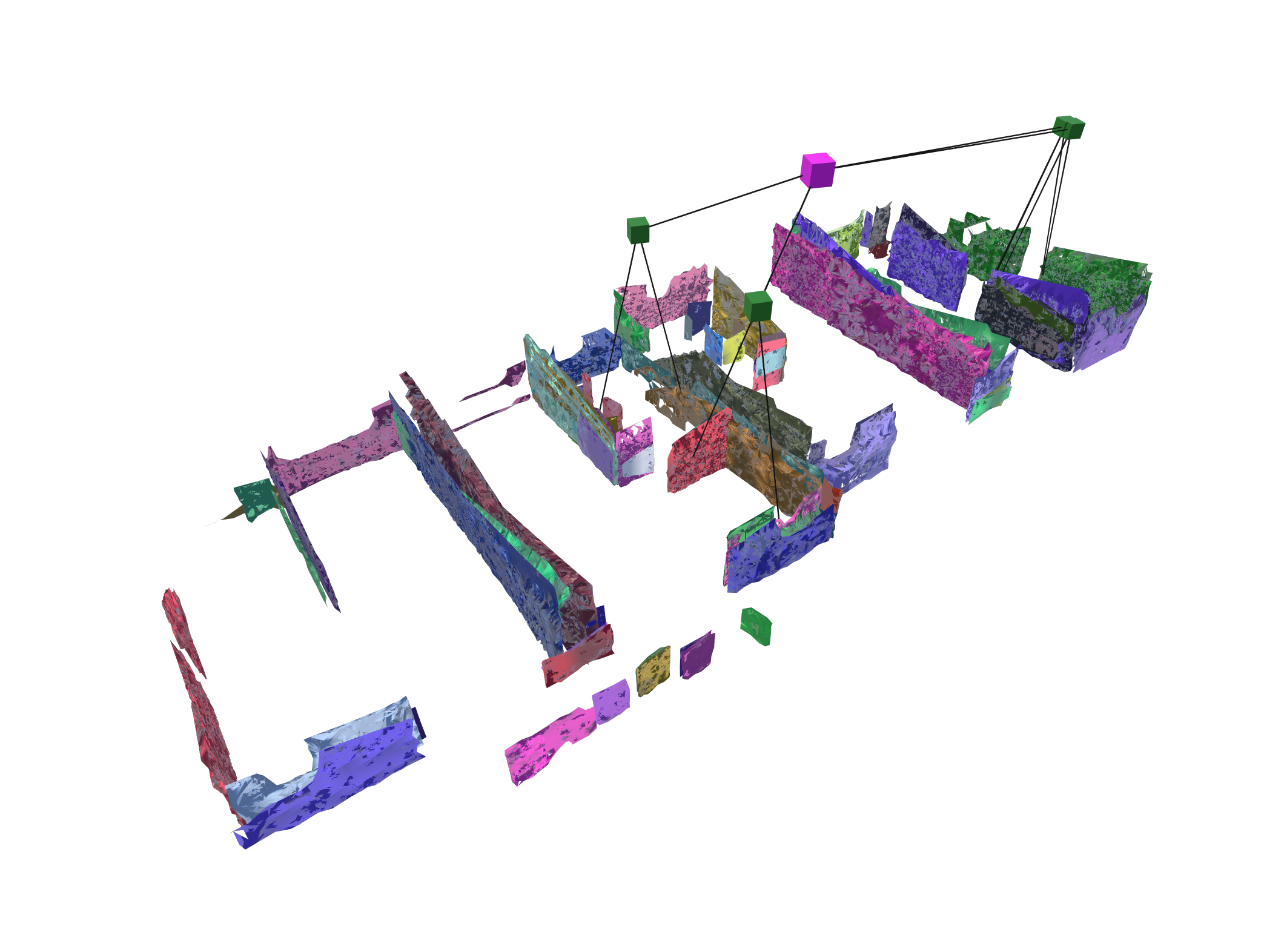}
    \caption{C2F1 S-Graphs}
    \vspace{0.3cm}
  \end{subfigure}
  \hfill
  \begin{subfigure}{0.3\textwidth}
    \centering
    \includegraphics[width=\linewidth]{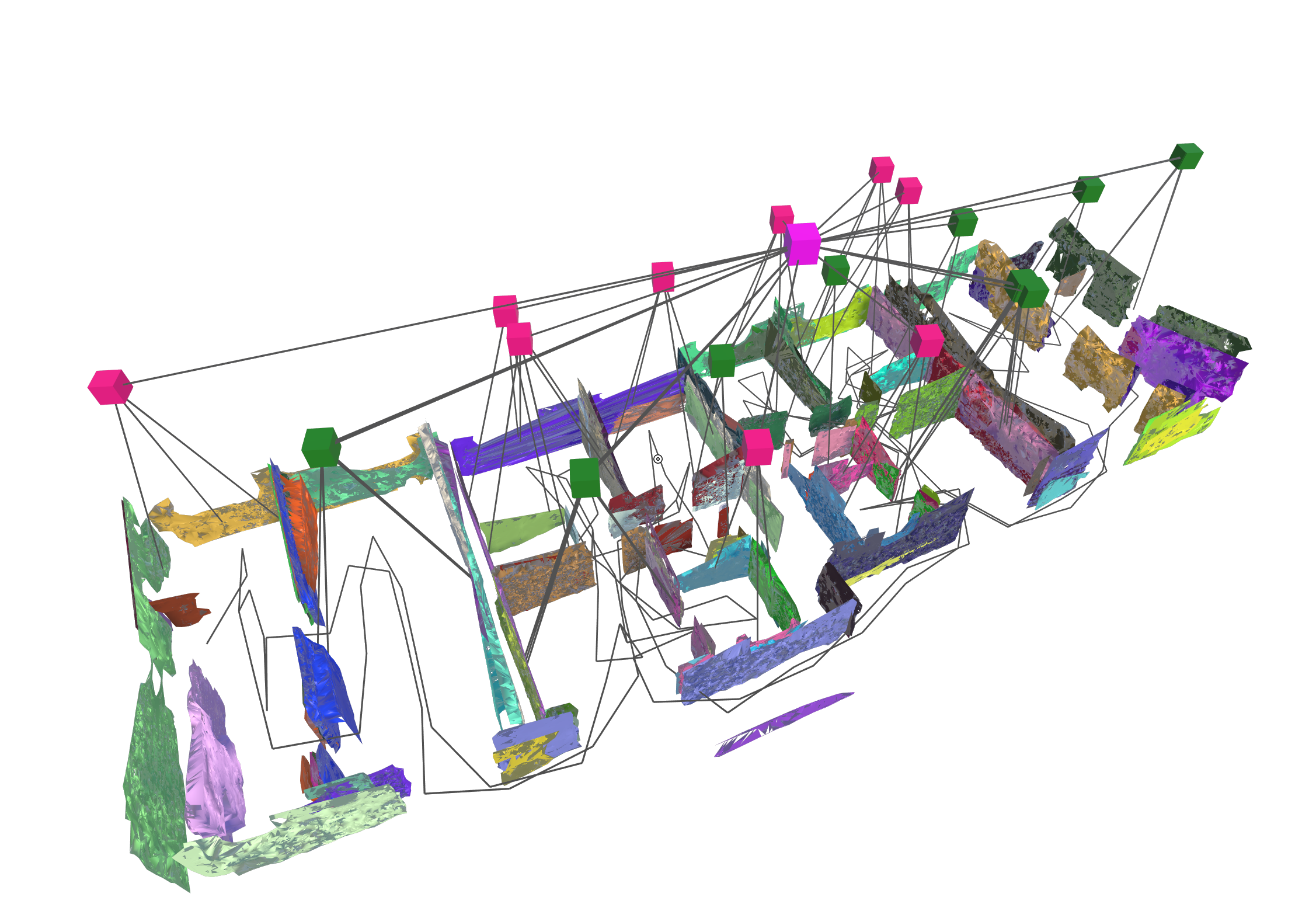}
    \caption{C2F2 S-Graphs}
    \vspace{0.3cm}
  \end{subfigure}
  \hfill
  \begin{subfigure}{0.3\textwidth}
    \centering
    \includegraphics[width=\linewidth]{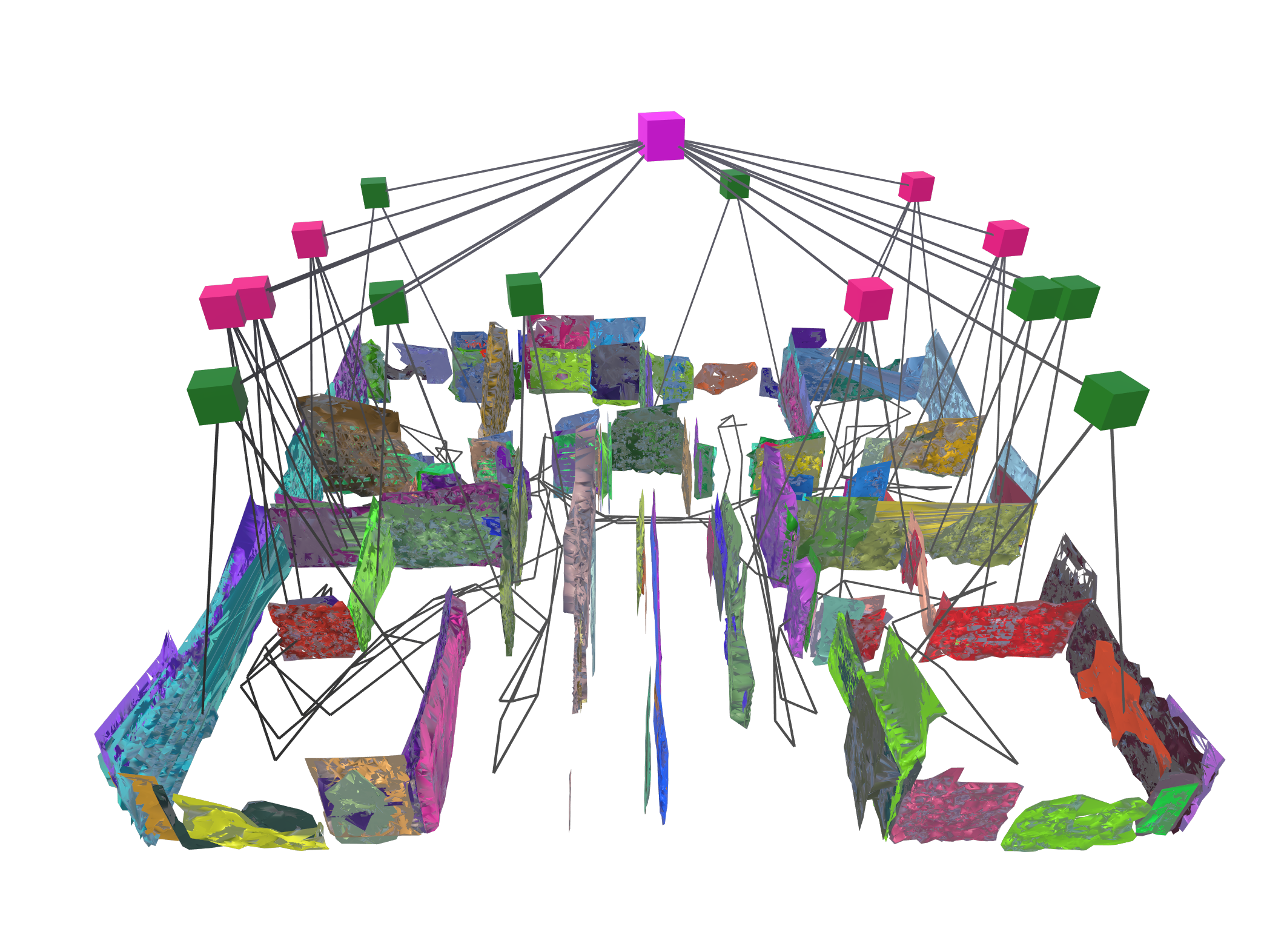}
    \caption{C3F2 S-Graphs}
    \vspace{0.3cm}
  \end{subfigure}
  \begin{subfigure}{0.3\textwidth}
    \centering
    \includegraphics[width=\linewidth]{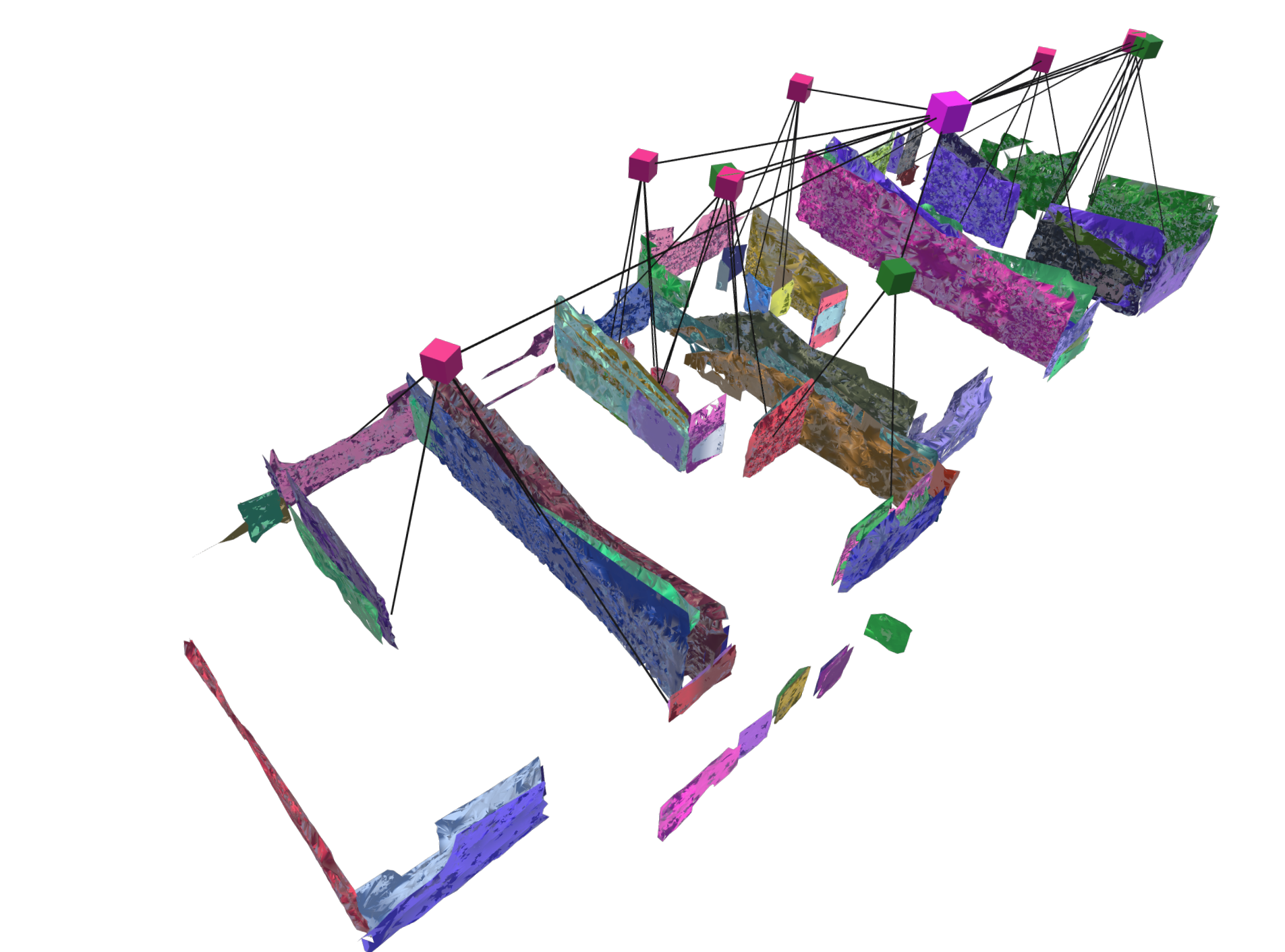}
    \caption{C2F1 Ours}
    \vspace{0.3cm}
  \end{subfigure}
  \hfill
  \begin{subfigure}{0.3\textwidth}
    \centering
    \includegraphics[width=\linewidth]{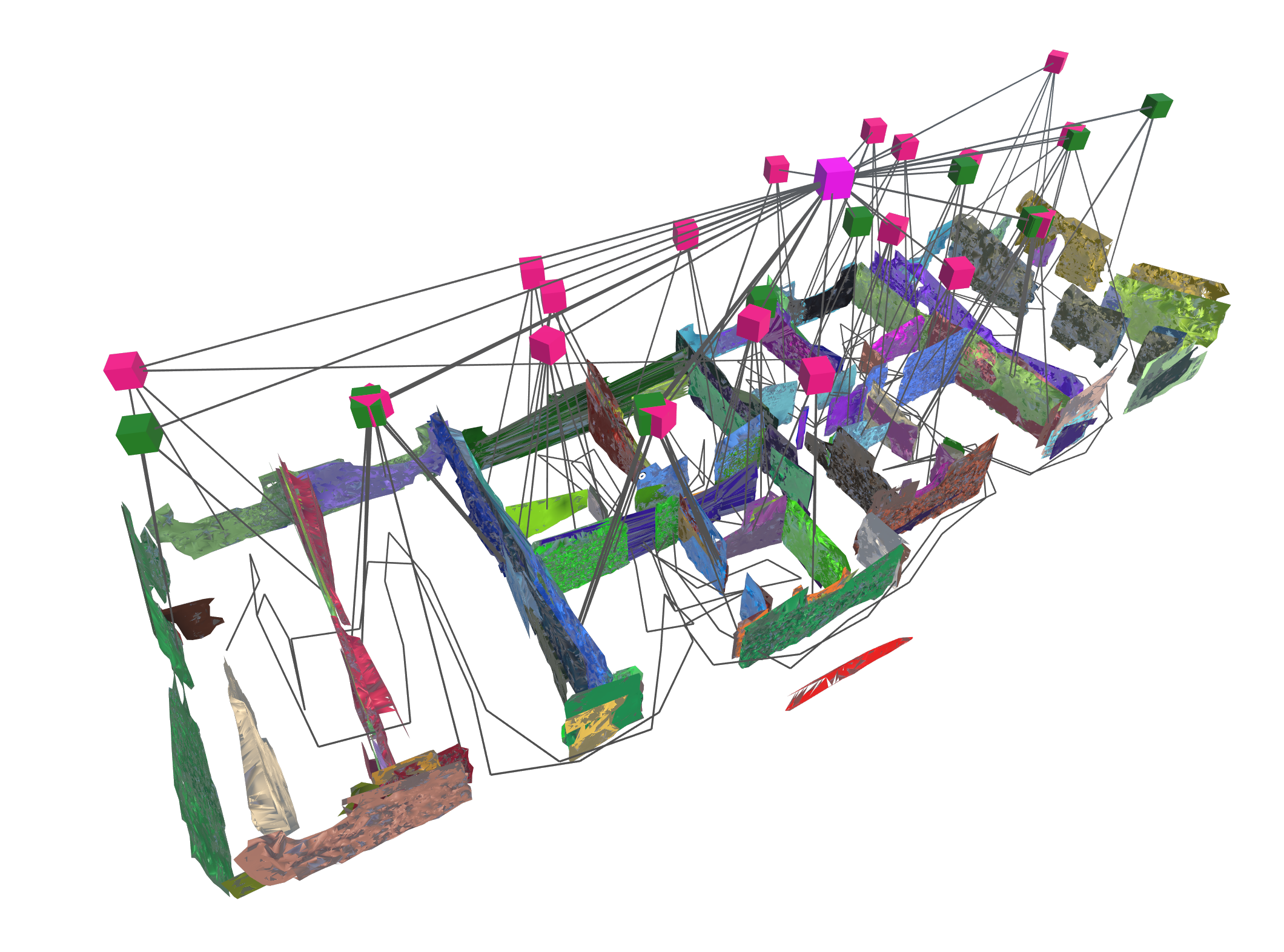}
    \caption{C2F2 Ours}
    \vspace{0.3cm}
  \end{subfigure}
    \vspace{0.3cm}
  \hfill
  \begin{subfigure}{0.3\textwidth}
    \centering
    \includegraphics[width=\linewidth]{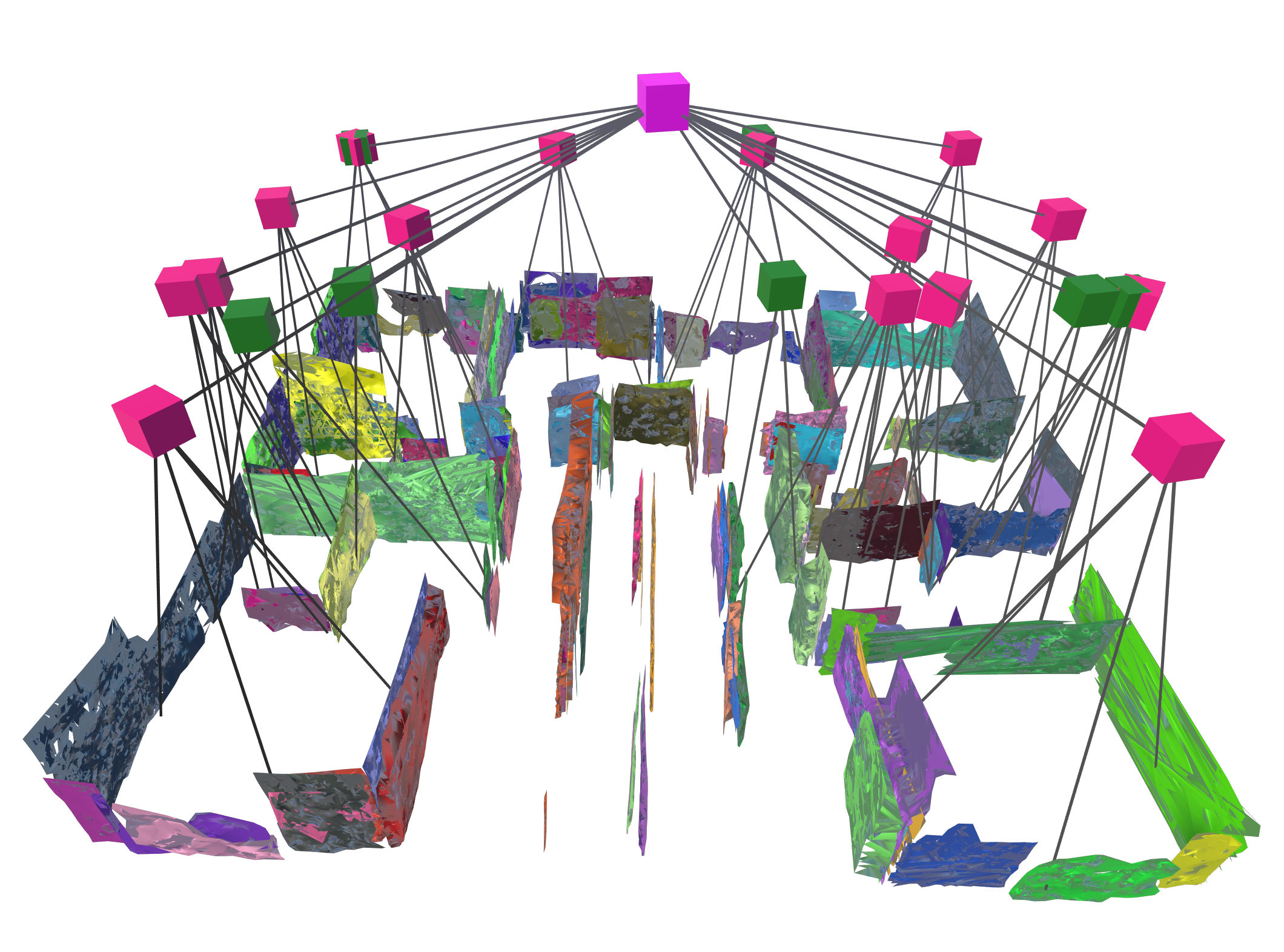}
    \caption{C3F2 Ours}
    \vspace{0.3cm}
  \end{subfigure}

  \caption{Qualitative results of room detection in three real-world scenarios (C2F1, C2F2, C3F2). Subfigures (a–c) show the baseline S-Graphs results, while (d–f) present the outcomes of HICS-SLAM, our proposed method. All visualizations are shown from the human perspective through XR.}
  \label{fig:qualitative_results}
\end{figure*}

Figure~\ref{fig:qualitative_results} demonstrates human-enhanced room detection across three representative scenarios (\textit{C2F1}, \textit{C2F2}, \textit{C3F2}) where automated classification differs from human spatial understanding.

In \textit{C2F1}, S-Graphs classifies the space as corridors while humans identify distinct enclosed rooms from the same geometric data. \textit{C2F2} shows partial room detection by S-Graphs, with humans recognizing additional rooms within the available point cloud. \textit{C3F2} illustrates how human operators successfully segment spaces into rooms despite automated algorithms struggling with the same geometric information. These cases highlight how human spatial reasoning can reinterpret identical geometric reconstructions to achieve superior semantic classification, directly supporting the quantitative results.

\subsection{Discussion}
The experiments demonstrate that integrating human corrections through the XR interface enhances the semantic completeness of the map while preserving the geometric accuracy of the baseline SLAM system. This balance highlights the potential of collaborative SLAM for real-world applications.

A critical aspect is that the effectiveness of the human intervention strongly depends on the accuracy of plane selection. If structural planes are chosen incorrectly, errors can propagate into the scene graph and reduce map reliability. To mitigate this, the interface asks users to confirm their selections before sending them to the SLAM system. Human operators contribute trusted inferences about spatial relationships and room boundaries, creating a collaborative framework where human situational awareness directly augments the robot's understanding.

This collaborative advantage becomes particularly evident in constrained environments, such as small rooms where robotic mobility is restricted, and during systematic identification failures, most notably in \textit{C2F1} (construction site) and \textit{C3F2} (residential building). These results highlight the importance of human cognitive input in bridging perceptual gaps for current autonomous mapping systems.

\textbf{Limitations:}
While the XR interactions are designed to be intuitive, factors including user training, experience, and spatial awareness may influence hand gesture tracking performance. The quality outcomes can be influenced by the operator, as inconsistent selection of structural planes can affect the accuracy of room annotations.

A practical consideration in deploying in XR/MR devices is their limited battery autonomy, which may restrict continuous operation during extended mapping sessions. However, this can be mitigated with scheduled recharging or by alternating devices among operators. Another aspect is that HoloLens 2 performs a short calibration when used by a new operator. While this adds some overhead, it ensures reliable gesture recognition and interaction quality, which is essential for collaborative use in real-world environments.

\section{Conclusion and Future Work}
\label{sec_conclusions}

This work introduces HICS-SLAM, a collaborative semantic SLAM framework using Extended Reality that bridges automated scene graph reconstruction with human spatial reasoning. The system creates an interactive shared space in a virtual world that represents the robot's situational awareness and enables collaborative semantic SLAM.

Our approach addresses the key limitation of autonomous SLAM systems that fail to segment and classify spatial boundaries properly. By enabling interactive room creation, the framework enhances incomplete semantic maps with structured spatial representations while maintaining geometric accuracy. This collaborative approach reduces post-processing requirements and allows real-time semantic enhancement without requiring complete remapping.

Future work will expand the supported operations to include editing and merging planes, correcting plane identifiers, and resolving duplicate structures. These capabilities are expected to improve both semantic accuracy and geometric precision, bringing the system closer to practical deployment. Additionally, comprehensive user studies will be conducted to validate the framework and guide improvements, ensuring its applicability across diverse deployment scenarios.

\bibliographystyle{IEEEtran}
\bibliography{root}

\end{document}